\pdfoutput=1

\documentclass[11pt]{article}

\usepackage{EMNLP2022}

\usepackage{times}
\usepackage{latexsym}
\usepackage{xspace}
\usepackage{textcomp}
\usepackage{booktabs}
\usepackage{multirow}
\usepackage{mathtools}
\usepackage{listings}

\definecolor{codegreen}{rgb}{0,0.6,0}
\definecolor{codegray}{rgb}{0.5,0.5,0.5}
\definecolor{codepurple}{rgb}{0.58,0,0.82}
\definecolor{backcolour}{rgb}{0.95,0.95,0.92} 

\lstdefinestyle{mystyle}{
    commentstyle=\color{codegreen},
    keywordstyle=\color{magenta},
    numberstyle=\tiny\color{codegray},
    stringstyle=\color{codepurple},
    basicstyle=\ttfamily\footnotesize,
    breakatwhitespace=false,         
    breaklines=true,                 
    captionpos=b,                  
    keepspaces=true,                 
    numbers=left,                    
    numbersep=5pt,                  
    showspaces=false,                
    showstringspaces=false,
    showtabs=false,                  
    tabsize=2
}
 
\newcommand{\spavg}{\textsc{P-SP}\xspace}

\newcommand{\spavgar}{\textsc{P-SP-ar}\xspace}
\newcommand{\spavgde}{\textsc{P-SP-de}\xspace}
\newcommand{\spavges}{\textsc{P-SP-es}\xspace}
\newcommand{\spavgfr}{\textsc{P-SP-fr}\xspace}
\newcommand{\spavgru}{\textsc{P-SP-ru}\xspace}
\newcommand{\spavgtr}{\textsc{P-SP-tr}\xspace}
\newcommand{\spavgzh}{\textsc{P-SP-zh}\xspace}

\newcommand{\spavgfull}{\textsc{Paragram-SP}\xspace}

\newcommand{\paragramphrase}{\textsc{paragram-phrase}\xspace}

\usepackage[T1]{fontenc}

\usepackage[utf8]{inputenc}

\usepackage{microtype}

\usepackage{inconsolata}

\title{Paraphrastic Representations at Scale}

\author{John Wieting$^1$, Kevin Gimpel$^2$, Graham Neubig$^3$, and Taylor Berg-Kirkpatrick$^4$ \\
  $^1$Google Research \\
  $^2$Toyota Technological Institute at Chicago, Chicago, IL, 60637, USA \\
  $^3$Carnegie Mellon University,
  Pittsburgh, PA, 15213, USA \\
  $^4$University of California San Diego,
  San Diego, CA, 92093, USA\\
  {\small \texttt{jwieting@alumni.cmu.edu}, \texttt{kgimpel@ttic.edu}, \texttt{gneubig@cs.cmu.edu}, \texttt{tberg@eng.ucsd.edu}}}

\date{}

\begin{document}
\maketitle
\begin{abstract}
We present a system that allows users to train their own state-of-the-art paraphrastic sentence representations in a variety of languages. We release trained models for English, Arabic, German, Spanish, French, Russian, Turkish, and Chinese. We train these models on large amounts of data, achieving significantly improved performance from our original papers on a suite of monolingual semantic similarity, cross-lingual semantic similarity, and bitext mining tasks. Moreover, the resulting models surpass all prior work on efficient unsupervised semantic textual similarity, even significantly outperforming supervised BERT-based models like Sentence-BERT~\cite{reimers-gurevych-2019-sentence}. Most importantly, our models are orders of magnitude faster than other strong similarity models and can be used on CPU with little difference in inference speed (even improved speed over GPU when using more CPU cores), making these models an attractive choice for users without access to GPUs or for use on embedded devices. Finally, we add significantly increased functionality to the code bases for training paraphrastic sentence models, easing their use for both inference and for training them for any desired language with parallel data. We also include code to automatically download and preprocess training data.\footnote{Code, including an easy to install PyPi package, released models including Hugging Face implementations, demo, and data are available at \url{https://github.com/jwieting/paraphrastic-representations-at-scale}.}
\end{abstract}

\section{Introduction}
Measuring sentence similarity~\cite{agirre-etal-2012-semeval} is an important task in natural language processing, and has found many uses including paraphrase detection~\cite{dolan-etal-2004-unsupervised}, bitext mining~\cite{schwenk-douze-2017-learning}, language modelling~\cite{khandelwal2019generalization}, question-answering~\cite{lewis-etal-2021-paq}, and as reward functions or evaluation metrics for language generation tasks~\cite{wieting-etal-2019-beyond}. 
Within this context, fast and light-weight methods are particularly useful as they make it easy to compute similarity over the ever-increasing volumes of web text available. For instance, we may want to mine a hundred million parallel sentences~\cite{schwenk-etal-2021-wikimatrix} or use a semantic similarity reward when fine-tuning language generation models on tens of millions of training examples. These tasks are much more feasible when using approaches that are fast, can be run on CPU, and use little RAM, allowing for increased batch size.

This need for fast inference is one motivation for using sentence embeddings. Sentence embeddings allow the search for similar sentences to be linear in the number of sentences, or even sub-linear when using highly optimized tools like \texttt{Faiss}~\cite{JDH17} that allow for efficient nearest neighbor search. This is contrast to models, like cross-attention models, which are quadratic during inference as they require both of the texts being compared as inputs. As we show in this paper, our {\it simple} and {\it interpretable} word-averaging sentence embedding models~\cite{wieting-16-full,wieting-gimpel-2018-paranmt,wieting-etal-2019-simple}, are orders of magnitude faster to compute than prior embedding approaches while simultaneously possessing significantly stronger performance on monolingual and cross-lingual semantic similarity tasks. Since we are simply averaging embeddings and have no neural architecture, any models based on neural architectures, especially large pretrained neural architectures which are increasingly used, will not be as fast as the models described in this paper. Lastly, we also show that this approach is competitive with LASER~\cite{artetxe-schwenk-2019-massively}, a state-of-the-art multilingual model, on mining bitext and has stronger performance on cross-lingual semantic similarity, while having  inference speeds that are twice as fast on GPU and orders of magnitude faster on CPU.

We make several contributions in this paper that go beyond our prior work. Firstly, we reformat the code to support training models on tens of millions of sentence pairs efficiently and with low RAM usage. Secondly, we train an English model on 25.85 million paraphrase pairs from ParaNMT~\cite{wieting-gimpel-2018-paranmt}, a paraphrase corpus we previously constructed automatically from bitext. We then train models directly on X-English bitext for Arabic, German, Spanish, French, Russian, Turkish, and Chinese, producing models that are able to distinguish both paraphrases in English and their respective languages as well as cross-lingual X-English paraphrases. Even though all models are able to model semantic similarity in English, we find that training on ParaNMT specifically leads to stronger models as it is easier to filter the data to remove noise and sentence pairs with little to no diversity. We refer to our models as \spavgfull, abbreviated as \spavg,\footnote{Our English model is \spavg, and the cross-lingual models are \spavgar, \spavgde, \spavges, \spavgfr, \spavgru, \spavgtr, and \spavgzh.}  referring to how the models are based on averaging subword units generated by \texttt{sentencepiece}~\cite{kudo-richardson-2018-sentencepiece}. We make all of these models available to the community for use on downstream tasks.

We also add functionality to our implementation. Besides the support for efficient, low-memory training on tens of million of sentence pairs described above, we add code to support (1) reading in a list of sentences and producing a saved \texttt{numpy} array of the sentence embeddings; (2) reading in a list of sentence pairs and producing cosine similarity scores; and (3) downloading and preprocessing evaluation data, bitext, and paraphrase data. For bitext and paraphrase data, we provide support for training using either text files or \texttt{HDF5} files.

Lastly, this paper contains new experiments showcasing the limits of these scaled-up models and detailed comparisons with prior work on a suite of semantic similarity tasks in a variety of languages. We release our code and models to the community in the hope that they will be found useful for research and applications, as well as using them as a base to build stronger, faster models covering more of the languages of the world.

\section{Related Work}
\subsection{English Semantic Similarity}
Our learning and evaluation setting is the same as that of our earlier work that seeks to learn paraphrastic sentence embeddings that can be used for downstream tasks~\cite{wieting-16-full,wieting-etal-2016-charagram,wieting-gimpel-2017-revisiting,wieting-etal-2017-learning,wieting-gimpel-2018-paranmt}. We trained models on noisy paraphrase pairs and evaluated them primarily on semantic textual similarity (STS) tasks. More recently, we made use of parallel bitext for training paraphrastic representations for other languages as well that are also able to model cross-lingual semantic similarity \cite{wieting-etal-2019-beyond,wieting-etal-2020-bilingual}. Prior work in learning general sentence embeddings has used  autoencoders \cite{SocherEtAl2011:PoolRAE,hill-etal-2016-learning}, 
encoder-decoder architectures \cite{kiros2015skip,gan-etal-2017-learning}, and other sources of supervision and learning frameworks \cite{le2014distributed,pham-etal-2015-jointly,arora2017simple,pagliardini2017unsupervised}.

For English semantic similarity, we compare to well known sentence embedding models such as InferSent \cite{conneau-etal-2017-supervised}, GenSen \cite{subramanian2018learning}, the Universal Sentence Encoder (USE) \cite{cer-etal-2018-universal}, as well as BERT \cite{devlin-etal-2019-bert}.\footnote{Note that in all experiments using BERT, including Sentence-BERT, the large, uncased version is used.} 
We use the pretrained BERT model in two ways to create a sentence embedding. The first way is to concatenate the hidden states for the CLS token in the last four layers. The second way is to concatenate the hidden states of all word tokens in the last four layers and mean pool these representations. Both methods result in a 4096 dimension embedding. We also compare to a more recently released model called Sentence-BERT \cite{reimers-gurevych-2019-sentence}. This model is similar to InferSent in that it is trained on natural language inference data (SNLI; \citealp{bowman-etal-2015-large}). However, instead of using pretrained word embeddings, they fine-tune BERT in a way to induce sentence embeddings.  Lastly, we also compare to the unsupervised version of SimCSE~\cite{gao-etal-2021-simcse}, which fine-tunes a pretrained encoder on contrastive pairs, where positive pairs are obtained by using dropout on a single input sentence.

\subsection{Cross-Lingual Semantic Similarity and Semantic Similarity in Non-English Languages}
Most previous work for cross-lingual representations has focused on models based on encoders from neural machine translation \cite{espana2017empirical,schwenk-douze-2017-learning,schwenk-2018-filtering} or deep architectures using contrastive losses \cite{gregoire-langlais-2018-extracting,guo-etal-2018-effective,chidambaram-etal-2019-learning}. Recently, other approaches using large Transformer~\cite{vaswani2017attention} have been proposed, trained on vast quantities of text~\cite{conneau-etal-2020-unsupervised,liu-etal-2020-multilingual-denoising,tran2020cross}. We primarily focus our comparison for these settings on LASER~\cite{artetxe-schwenk-2019-massively}, a model trained for semantic similarity across more than 100 languages. Their model uses an LSTM encoder-decoder trained on hundreds of millions of parallel sentences. They achieve state-of-the-art performance on a variety of multilingual sentence embeddings tasks including bitext mining. We also compare to LaBSE~\cite{feng-etal-2022-language}, a contrastive model trained on six billion parallel pairs across languages and was also trained on monolingual text using a masked language modelling objective.

\section{Methods}
We first describe our objective function and then describe our encoder. 

\paragraph{Training.}
The training data consists of a sequence of parallel sentence pairs $(s_i, t_i)$ in source and target languages respectively. Note that for training our English model, the source and target languages are both English as we are able to make use of an existing paraphrase corpus. For each sentence pair, we randomly choose a \emph{negative} target sentence $t_i'$ during training 
that is not a translation or paraphrase of $s_i$. Our objective is to have source and target sentences be more similar than source and negative target examples by a margin $\delta$:
\begin{align}
\min_{\theta_\textrm{src}, \theta_\textrm{tgt}} \sum_i\Big[\delta - &f_{\theta}(s_i,t_i) + f_{\theta}(s_i,t_i'))\Big]_+
\label{eq:obj}
\end{align}
where the similarity function is defined as:
\begin{align}
f_\theta(s,t) = \textrm{cos}\Big(&g(s;\theta_\textrm{src}), g(t; \theta_\textrm{tgt})\Big)
\end{align} 
\noindent where $g$ is the sentence encoder with parameters for each language $\theta = (\theta_\textrm{src}, \theta_\textrm{tgt})$.
To select $t_i'$ we choose the most similar sentence in some set according to the current model parameters, i.e., the one with the highest cosine similarity. We found we could achieve the strongest performance by tying all parameters together for each language, more precisely, $\theta_\textrm{src}$ and $\theta_\textrm{tgt}$ are the same.

\paragraph{Negative Sampling.} Negative examples are selected from the sentences in the batch from the opposing language when training with bitext and from any sentence in the batch when using paraphrase data. In all cases, we choose the negative example with the highest cosine similarity to the given sentence $s$, ensuring that the negative is not in fact paired with $s$ in the batch. 
To select even more difficult negative examples that aid training, we use the {\it mega-batching} procedure of \citet{wieting-gimpel-2018-paranmt}, which aggregates $M$ mini-batches to create one ``mega-batch'' and selects negative examples from this mega-batch. Once each pair in the mega-batch has a negative example, the mega-batch is split back up into $M$ mini-batches for training. Additionally, we anneal the mega-batch size by slowly increasing it during training. This yields improved performance by a significant margin. 

\paragraph{Encoder.} Our sentence encoder $g$ simply averages the embeddings of subword units generated by \texttt{sentencepiece}~\cite{kudo-richardson-2018-sentencepiece};
we refer to our model as \spavgfull, abbreviated as \spavg. This means that the sentence piece embeddings themselves are the only learned parameters of this model.

\section{Code and Usage} \label{sec:code}
We added a number of features to the code base to improve performance and make it easier to use. First, we added code to support easier inference. Examples of using the code programmatically to embed sentences and score sentence pairs (using cosine similarity) are shown in Figure~\ref{fig:program}. 

\begin{figure}[h!]
\begin{lstlisting}[language=Python]
from models import load_model

text1 = 'This is a test.'
text2 = 'This is another test.'

# Load English paraphrase model
model_name = 'paraphrase-at-scale/model.para.lc.100.pt'
sp_model = 'paraphrase-at-scale/paranmt.model'

model, _ = load_model(model_name=model_name, sp_model=sp_model)

# Obtain sentence embedding
embeddings = model.embed_raw_text([text1, text2]) # 2D numpy array of sentence embeddings
cosine_scores = model.score_raw_text([(text1, text2)]) # list of cosine scores
\end{lstlisting} 
\caption{Usage example of programmatically loading one of our pretrained models and obtaining sentence embeddings and scores for two sentences.}
\label{fig:program}
\end{figure}

Our code base also supports functionality that allows one to read in a list of sentences and produce a saved \texttt{numpy} array of the sentence embeddings. We also included functionality that allows one to read in a list of sentence pairs and produce the sentence pairs along with their cosine similarity scores in an output file. These scripts allow our models to be used without any programming for the two most common use cases: embedding sentences and scoring sentence pairs. Examples of their usage with a trained model are shown in Figure~\ref{fig:bash2}.

\begin{figure}[h!]
\begin{lstlisting}[language=bash]
python -u embed_sentences.py --sentence-file paraphrase-at-scale/example-sentences.txt --load-file paraphrase-at-scale/model.para.lc.100.pt --output-file sentence_embeds.np

python score_sentence_pairs.py --sentence-pair-file paraphrase-at-scale/example-sentences-pairs.txt --load-file paraphrase-at-scale/model.para.lc.100.pt
\end{lstlisting}
\caption{Usage examples to embed sentences and score sentence pairs. The first command is a usage example of scoring a list of sentence pairs. The file \texttt{example-sentences-pairs.txt} contains a list of sentences, one per line. The output of the script is a saved \texttt{numpy} array of sentence embeddings in the same order of the input sentences. The second command is a usage example of scoring a list of sentence pairs. The file \texttt{example-sentences-pairs.txt} contains pairs of tab-separated sentences, one per line. The output of the script is a text file containing the tab separated list of sentences along with their cosine scores in the same order of the input sentences.}
\label{fig:bash2}
\end{figure}

Secondly, we added a training mode using \texttt{HDF5}\footnote{\url{https://docs.h5py.org/en/stable/}} format, allowing training data to remain on disk during training. This leads to a significant reduction in RAM usage during training, which is especially true when using more than 10 million training examples. Efficient training can now be done on CPU only using only a few gigabytes of RAM.

\begin{figure}[h!]
\begin{lstlisting}[language=bash]
cd preprocess/bilingual && bash do_all.sh fr-es-de
cd ../..
cd preprocess/paranmt && bash do_all.sh 0.4 1.0 0.7
\end{lstlisting}
\caption{Usage examples to download and preprocess bilingual and ParaNMT data. The first command downloads and preprocesses (filters, trains \texttt{sentencepiece} models, tokenizes if language is \texttt{zh}, converts files to \texttt{hdf5} format) \texttt{en-X} bilingual data. The third command downloads and preprocesses ParaNMT data. The arguments are used to filter the data (semantic similarity scores between 0.4 and 1.0 and trigram overlap below 0.7, which have been used in prior papers when generating training data for paraphrase generation~\cite{iyyer-etal-2018-adversarial,krishna-etal-2020-reformulating}).}
\label{fig:bash3}
\end{figure}

Lastly, we also added code for preprocessing data, including scripts to download and evaluate on the STS data (English, non-English, and cross-lingual), as well as code to download and process bitext and ParaNMT automatically. For bitext, our scripts download the data, filter the data by length,\footnote{We remove sentences with the number of tokens (untokenized) smaller than 3 or greater than 100.} lowercase, remove duplicates, train a \texttt{sentencepiece} model, encode the data with the \texttt{sentencepiece} model, shuffle the data, and process the data into \texttt{HDF5} format for efficient use. For ParaNMT, our scripts download the data, use a language classifier to filter out non-English sentences\footnote{\url{https://fasttext.cc}}~\cite{joulin-etal-2017-bag}, filter the data by paraphrase score, trigram overlap, and length,\footnote{We remove sentences with the number of tokens (untokenized) smaller than 5 or greater than 40.} train a \texttt{sentencepiece} model, encode the data with the \texttt{sentencepiece} model, and process the data into \texttt{HDF5} format. Examples are shown in Figure~\ref{fig:bash3}.

\section{Experiments}
\subsection{Experimental Setup} \label{sec:expt}

\paragraph{Data.}

\begin{table}
\begin{center}
\small
\setlength{\tabcolsep}{1.4pt}
\begin{tabular}{ cccccccc } 
 \toprule
\texttt{en} &  \texttt{ar} &  \texttt{de} &  \texttt{es} & \texttt{fr} & \texttt{ru} & \texttt{tr} &  \texttt{zh} \\
\toprule
25.85M & 8.23M & 6.47M & 6.75M & 6.46M & 9.09M & 5.12M & 4.18M\\
\bottomrule
\end{tabular}
\end{center}
\caption{The number of sentence pairs used to train our models. For English, the data is ParaNMT, and for the other languages, the data is a collection of bitext detailed in Section~\ref{sec:expt}. 
}
\label{tab:data}
\end{table}

For our English model, we train on selected sentence pairs from ParaNMT~\cite{wieting-gimpel-2018-paranmt}. We filter the corpus by only including sentence pairs where the paraphrase score for the two sentences is $\geq 0.4$. We additionally filtered sentence pairs by their trigram overlap~\citep{wieting-etal-2017-learning}, which is calculated by counting trigrams in the two sentences, and then dividing the number of shared trigrams by the total number in the sentence with fewer tokens. We only include sentence pairs where the trigram overlap score is $\leq 0.7$. The paraphrase score is calculated by averaging \paragramphrase embeddings~\cite{wieting-16-full} for the two sentences in each pair and then computing their cosine similarity. The purpose of the lower threshold is to remove noise while the higher threshold is meant to remove paraphrases that are too similar.

Our training data is a mixture of Open Subtitles 2018\footnote{\url{http://opus.nlpl.eu/OpenSubtitles.php}} \cite{lison-tiedemann-2016-opensubtitles2016}, Tanzil corpus\footnote{\url{http://opus.nlpl.eu/Tanzil.php}} \cite{tiedemann-2012-parallel}, Europarl\footnote{\url{http://opus.nlpl.eu/Europarl.php}} for Spanish, Global Voices\footnote{\url{https://opus.nlpl.eu/GlobalVoices.php}} \cite{tiedemann-2012-parallel}, and the MultiUN corpus.\footnote{\url{http://opus.nlpl.eu/MultiUN.php}} We follow the same distribution for our languages of interest across data sources as \citet{artetxe-schwenk-2019-massively} for a fair comparison. One exception, though, is we do not include training data from Tatoeba\footnote{\url{https://opus.nlpl.eu/Tatoeba.php}} \cite{tiedemann-2012-parallel} as they do, since this domain is also in the bitext mining evaluation set. The amount of data used to train each of our models is shown in Table~\ref{tab:data}.

\paragraph{Hyperparameters.}
For all models, we fix the batch size to 128, margin $\delta$ to 0.4, and the annealing rate to 150.\footnote{Annealing rate is the number of minibatches that are processed before the megabatch size is increased by 1.} We set the size of the \texttt{sentencepiece} vocabulary to 50,000, using a shared vocabulary for the models trained on bitext. If a word is not in vocabulary, we simply exclude it, unless the text only consists of unknown words in which case we use a single unknown-word token. We optimize our models using Adam~\cite{kingma2014adam} with a learning rate of 0.001 and train models for 25 epochs.

For training on the bilingual corpora, we tune each model on the 250 example 2017 English STS task~\cite{cer-etal-2017-semeval}. We vary dropout on the embeddings over $\{0, 0.1, 0.3\}$ and the mega-batch size $M$ over $\{60, 100, 140\}$.

For training on ParaNMT, we fix the hyperparameters in our model due to the increased data size making tuning more expensive. We use a mega-batch size $M$ of 100 and set the dropout on the embeddings to 0.0. 

\subsection{Evaluation}

\begin{table*}
\centering
\small
\begin{tabular} { lccccc|c}
\toprule
\multirow{2}{*}{Model}  & \multicolumn{6}{c}{Semantic Textual Similarity (STS)}\\
& 2012 & 2013 & 2014 & 2015 & 2016 & \bf Avg. \\
\hline
BERT (CLS) & 33.2 & 29.6 & 34.3 & 45.1 & 48.4 & 38.1\\
BERT (Mean) & 48.8 & 46.5 & 54.0 & 59.2 & 63.4 & 54.4\\
InferSent & 61.1 & 51.4 & 68.1 & 70.9 & 70.7 & 64.4\\
GenSen & 60.7 & 50.8 & 64.1 & 73.3 & 66.0 & 63.0\\
USE & 61.4 & 59.0 & 70.6 & 74.3 & 73.9 & 67.8\\
Sentence-BERT & 66.9 & 63.2 & 74.2 & 77.3 & 72.8 & 70.9\\
LASER & 63.1 & 47.0 & 67.7 & 74.9 & 71.9 & 64.9\\
\spavg & \bf 68.7 & \bf 64.7 & \bf 78.1 & \bf 81.4 & \bf 80.0 & \bf 74.6\\
\midrule 
Sentence-BERT & 71.0 & \bf 76.5 & 73.2 & 79.1 & 74.3 & 74.8\\
\spavg & \bf 71.2 & \bf 76.5 & \bf 74.6 & \bf 83.0 & \bf 79.1 & \bf 76.9\\
\bottomrule
\end{tabular}
\caption{\label{table:english-sts} Results of our models and models from prior work on English STS. In the first part of the table, we show results, measured in Pearson's $r \times 100$, for each year of the STS tasks 2012-2016 as well as the average performance across all years. In the second part, we evaluate based on the Spearman's $\rho \times 100$ of the concatenation of the datasets of each year with the 2013 SMT dataset removed following~\cite{reimers-gurevych-2019-sentence}.}
\end{table*}

We evaluate sentence embeddings using the 
SemEval semantic textual similarity (STS) tasks from 2012 to 2016~\cite{agirre-etal-2012-semeval,agirre-etal-2013-sem,agirre-etal-2014-semeval,agirre-etal-2015-semeval,agirre-etal-2016-semeval} as was done initially for sentence embeddings in~\cite{wieting-16-full}.
Given two sentences, the aim of the STS tasks is to predict their similarity on a 0-5 scale, where 0 indicates the sentences are on different topics and 5 means they are completely equivalent. As our test set, we report the average Pearson's $r$ over each year of the STS tasks from 2012-2016 as is convention.

Most work evaluating accuracy on STS tasks has averaged the Pearson's $r$ over each individual dataset for each year of the STS competition. However, \citet{reimers-gurevych-2019-sentence} computed Spearman's $\rho$ over concatenated datasets for each year of the STS competition. To be consistent with previous work, we re-ran their model and calculated results using the standard method, and thus our results are not the same as those reported \citet{reimers-gurevych-2019-sentence}. However, we also include results using their approach for completeness. One other difference between these two ways of calculating the results is the inclusion of the SMT dataset of the 2013 task, which we also exclude when replicating the approach in \citet{reimers-gurevych-2019-sentence}.

For cross-lingual semantic similarity and semantic similarity in non-English languages, we evaluate on the STS tasks from SemEval 2017. This evaluation contains Arabic-Arabic, Arabic-English, Spanish-Spanish, Spanish-English, and Turkish-English datasets. The datasets were created by translating one or both pairs of an English STS pair into Arabic (\texttt{ar}), Spanish (\texttt{es}), or Turkish (\texttt{tr}). Following convention, we report results with Pearson's $r$ for all systems, but also include results in Spearman's $\rho$ for LASER, LaBSE, and \spavg.

We also evaluate on the Tatoeba bitext mining task introduced by \citet{artetxe-schwenk-2019-massively}. The dataset consists of up to 1,000 English-aligned sentence pairs
for over 100 languages. The aim of the task is to find the nearest neighbor for each sentence in the other language according to cosine similarity. Performance is measured by computing the error rate.

\section{Results}

\begin{table*}
\centering
\small
\begin{tabular} { lcccccccccccc }
\toprule
Model & Dim. & \multicolumn{2}{c}{\texttt{ar-ar}} & \multicolumn{2}{c}{\texttt{ar-en}} & \multicolumn{2}{c}{\texttt{es-es}} & \multicolumn{2}{c}{\texttt{es-en}} & \multicolumn{2}{c}{\texttt{tr-en}}\\
& & $r$ & $\rho$ & $r$ & $\rho$ & $r$ & $\rho$ & $r$ & $\rho$ & $r$ & $\rho$ \\
\midrule
LASER & 1024 & 69.3 & 68.8 & 65.5 & 66.5 & 79.7 & 79.7 & 59.7 & 58.0 & 72.0 & 72.1 \\
LaBSE & 768 & 68.6 & 69.1 & 72.2 & 74.5 & 79.5 & 80.8 & 65.5 & 65.7 & 72.9 & 72.1 \\
\citet{espana2017empirical} & 2048 & 59 & - & 44 & - & 78  & - & 49 & - & 76 & -\\
\citet{chidambaram-etal-2019-learning} & 512 & - & - & - & - & 64.2 & - & 58.7 & - & - & -\\
\midrule
2017 STS 1st Place & - & 75.4 & - & 74.9& - & 85.6& - & \bf 83.0& - & 77.1 & -\\
2017 STS 2nd Place & - & 75.4 & - & 71.3 & - & 85.0 & - & 81.3 & - & 74.2 & -\\
2017 STS 3rd Place & - & 74.6 & - & 70.0 & - & 84.9 & - & 79.1 & - & 73.6 & -\\
\midrule
\spavg & 1024 & \bf 76.2 & \bf 76.7 & \bf 78.3 & \bf 78.4 & \bf 85.8 & \bf 85.6 & 78.4 & \bf77.8 & \bf 79.2 & \bf 79.5 \\
\bottomrule
\end{tabular}
\caption{\label{table:multi} Comparison of our models with those in the literature on non-English and cross-lingual STS. We also include the top 3 systems for each dataset from the SemEval 2017 STS shared task. 
Performance is measured in Pearson's $r$ $\times 100$. We also include results in Spearman's $\rho$ $\times 100$ for LASER, LaBSE, and \spavg.
}\label{table:crosslingual-sts}
\end{table*}

\paragraph{English Semantic Similarity.}
The results for our English semantic similarity evaluation are shown in Table~\ref{table:english-sts}. Our \spavg model has the best performance across each year of the task, significantly outperforming all prior work. We outperform methods that use large pre-trained models including Sentence-BERT which is supervised, as it is trained on NLI data~\cite{bowman-etal-2015-large}.

We also include results from SimCSE~\cite{gao-etal-2021-simcse}. We compare to the unsupervised version, since our model is also unsupervised. We evaluate using the Spearman's $\rho$ of the concatenation of the datasets for each year, and find our average performance over the 2012-2016 datasets to be 76.9, compared to 77.4 and 77.9 for the RoBERTa-base~\cite{liu2019roberta} and RoBERTa-large versions of SimCSE. While our performance is slightly lower, we note that they tune their model on the dev set of the STS Benchmark~\cite{cer-etal-2017-semeval}, which contains a subset of the data from STS tasks which we use for evaluation. Therefore, they are tuning on a subset of the evaluation data, and it is unclear how tuning on this test data affects model performance.

\paragraph{Cross-Lingual Semantic Similarity.}
The results for the non-English and cross-lingual semantic similarity evaluation are shown in Table~\ref{table:crosslingual-sts}. From the results, our model again outperforms all prior work using sentence embeddings. The only systems that have better performance are the top (non-embedding based) systems from SemEval 2017 for Spanish-English.\footnote{The top systems for this task used supervision and relied on state-of-the-art translation models to first translate the non-English sentences to English.}

\paragraph{Bitext Mining.}

\begin{table}
\centering
\small
\setlength{\tabcolsep}{1.5pt}

\begin{tabular}{ lccccccc } 
 \toprule
Language & LASER & XLM-R & mBART & CRISS & LaBSE & \spavg\\
\toprule
\texttt{ar} & \bf 7.8 & 52.5 & 61.0 & 22.0 & 9.1 & 8.8\\
\texttt{de} & 1.0 & 11.1 & 13.2 & 2.0 & \bf 0.7 & 1.5\\
\texttt{es} & 2.1 & 24.3 & 39.6 & 3.7 & \bf 1.6 & 2.4\\
\texttt{fr} & 4.3 & 26.3 & 39.6 & 7.3 & \bf 4.0 & 5.4\\
\texttt{ru} & 5.9 & 25.9 & 31.6 & 9.7 & \bf 4.7 & 5.6\\
\texttt{tr} & 2.6 & 34.3 & 48.8 & 7.1 & 1.6 & \bf 1.4\\
\midrule
\bf Avg. & 4.0 & 29.1 & 39.0 & 8.6 & \bf 3.6 & 4.2\\
\bottomrule
\end{tabular}
\caption{Results on the Tatoeba bitext mining task \cite{artetxe-schwenk-2019-massively}. Results are measured in error rate $\times 100$.}
\label{tab:mining}
\end{table}

The results on the Tatoeba bitext mining task from \citet{artetxe-schwenk-2019-massively} are shown in Table~\ref{tab:mining}. The results show that our embeddings are competitive, but have slightly higher error rates than LASER. The models are so close that the difference in error rate for the two models across the 6 evaluations is 0.2, corresponding to a difference of about 2 mismatched sentence pairs per dataset. LaBSE performs a bit better, but was trained on much more data then both LASER and our method. We also compare to mBART, XLM-R, and CRISS.\footnote{Results are copied from~\cite{tran2020cross}.}

This bitext mining result is in contrast to the results on cross-lingual semantic similarity, suggesting that our embeddings account for a less literal semantic similarity, making them more adept at detecting paraphrases but slightly weaker at identifying translations. It is also worth noting that LASER was trained on Tatoeba data outside the test sets, which could also account for some of the slight improvement over our model.

\section{Speed Analysis}

\begin{table}[h!]
\centering
\small
\begin{tabular} { lcc}
\toprule
Model & GPU & CPU \\
\midrule
\spavg & \bf 13,863 & \bf 12,776 \\
LASER & 6,033 & 26 \\
Sentence-Bert & 288 & 2 \\
InferSent & 4,445 & 16 \\
\bottomrule
\end{tabular}
\caption{\label{table:bucc} Speed as measured in sentences/second on both GPU (Nvidia 1080 TI) and CPU (single core).
}
\label{tab:speed}
\end{table}

\noindent We analyze the speed of our models as well as selected popular sentence embedding models from prior work. To evaluate inference speed, we measure the time required to embed 120,000 sentences from the Toronto Book Corpus~\cite{Zhu_2015_ICCV}. Preprocessing of sentences is not factored into the timing, and each method sorts the sentences by length prior to computing the embeddings to reduce padding and extra computation. We use a batch size of 64 for each model. The number of sentences embedded per second is shown in Table~\ref{tab:speed}.

From the results, we see that our model is easily the fastest on GPU, sometimes by an order of magnitude. Interestingly, using a single core of CPU, we achieve similar speeds to inference on GPU, which is not the case for any other model. Moreover, we repeated the experiment, this time using 32 cores and achieved a speed of 15,316 sentences/second. This is even faster than when using a GPU and indicates that our model can effectively be used at scale when GPUs are not available. It also suggests our model would be appropriate for use on embedded devices.

\section{Conclusion}
In this paper, we present a system for the learning and inference of paraphrastic sentence embeddings in any language for which there is paraphrase or bilingual parallel data. Additionally, we release our trained sentence embedding models in English, as well as Arabic, German, Spanish, French, Russian, Turkish, and Chinese. These models are trained on tens of million of sentence pairs resulting in models that achieve state-of-the-art performance on unsupervised English semantic similarity and are state-of-the-art or competitive on non-English semantic similarity, cross-lingual semantic similarity, and bitext mining. 

Moreover, our models are significantly faster than prior work owing to their simple architecture. They can also be run on CPU with little to no loss in speed from running them on GPU----something that no strong models from prior work are able to do. Lastly, we release our code that has been modified to make training and inference easier, with support for training on large corpora, preprocessing paraphrase and bilingual corpora and evaluation data, as well as scripts for easy inference that can generate embeddings or semantic similarity scores for sentences supplied in a text file.

\bibliographystyle{acl_natbib}
\bibliography{emnlp2022}


\end{document}